# Cascaded Light Propagation Volumes using Spherical Radial Basis Functions

Ludovic Silvestre
ludovic.silvestre@tecnico.ulisboa.pt

João Pereira
jap@inesc-id.pt

Instituto Superior Técnico/INESC-ID

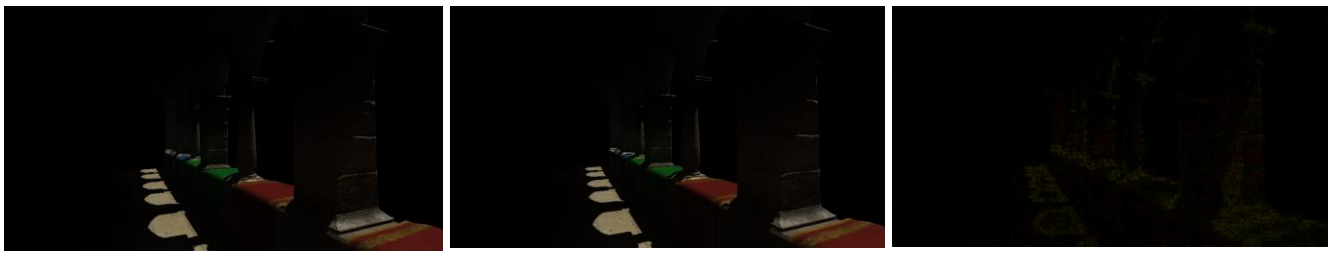

**Figure 1:** *The best case scenario for the Sponza Atrium with indirect lighting using a 2-band SH (left), a SRBF with 8 coefficients (middle) and the difference between both (right). Note there is no other lighting besides direct lighting and indirect lighting from the cascaded ligh propagation volumes (no ambient color or screen-space ambient occlusion).*

## Abstract

This paper introduces a contribution made to one of the newest methods for simulating indirect lighting in dynamic scenes, the cascaded light propagation volumes. Our contribution consists on using Spherical Radial Basis Functions instead of Spherical Harmonics, since the first achieves much better results when many coefficients are used. We explain how to integrate the Spherical Radial Basis Functions with the cascaded light propagation volumes, and evaluate our technique against the same implementation, but with Spherical harmonics.

**Keywords:** light propagation volumes, spherical radial basis functions

## 1. Introduction

Since the 80's, the simulation of indirect lighting has been one of the main challenges when creating virtual worlds. This affirmation is easily verifiable if we look at the number of discovered methods to solve this problem and at the diversity of the approaches taken by those methods. This is completely normal, considering that the simulation of indirect lighting is computationally expensive, due to the lighting equation's complexity. Nonetheless, due to hardware advances, the use of pre-computation [Sloan et. al. 2002] and quality trade-offs [Ritschel et. al. 2009] [Figueiredo et al. 2010], some of these methods achieves interactive or real-time performance.

In this work, we are analyzing one of the newest methods, the cascaded light propagation volumes (CLPV) [Kaplanyan et. al. 2010]. This method consists on injecting the intensity of indirect light sources into several volumes and propagating that intensity until the result converges. The result of this propagation represents the indirect lighting applied to the objects in those volumes. We choose this method because it is efficient and fast, it does not need any pre-computation and it is highly parallelizable. Those reasons not only enable the use of this method on video game consoles, but also allow simulating indirect lighting in completely dynamic environments, like destructible environments.

Our contribution to this method consists on using a different data type to store the indirect lighting intensity. The original method uses 2-band spherical harmonics (SH), because it is a computationally inexpensive data type, while still giving a satisfactory result. We decided to implement this method using spherical radial basis functions (SRBF), considering that this data type gives a better result than SH when both uses many coefficients [Tsai et. al. 2006]. Then, it is our objective to observe the SRBF's behavior when using a restricted number of coefficients. We expect to verify a small improvement on high frequency surfaces and in environments with a lot of blocking objects.

We will begin our document by introducing the previous works that our contribution was based upon (Section 2), followed by our contribution, which consist on explaining how the SRBF works for the CLPV (Section 3) and how they can be implemented using a GPU programming language (shader language, not general programming language). Then, we will present our qualitative evaluation (Section 4), followed by some details about our implementation (Section 5), our results (Section 6) and conclusions (Section 7).

## 2. Previous works

Our contribution is based upon two distinct methods for indirect lighting, but both share a common goal: simulating indirect lighting in interactive, real-time virtual worlds. We already mentioned the first method in the introduction, which is the cascaded light propagation volumes [Kaplanyan et. al. 2010] and is a great method to simulate indirect lighting for completely dynamic worlds. The second method is the pre-computed radiance transfer [Sloan et. al. 2002] (PRT), which is one of the most used methods for games, since it achieves a great result without using too much rendering time. A particular version of this method drawn our attention, which is PRT using SRBF instead of SH [Tsai et. al. 2006]. The result achieved with a high number of coefficients is impressive, but the most interesting part is the difference between the 3 tested data types: SH, SRBF and wavelets. Obviously, the SRBF gave a better result, being the reason why we choose this data type to combine with the CLPV.

Let us start by briefly explaining how the CLPV works. The CLPV uses two components to simulate indirect lighting and shadowing, one being cascaded light volumes and the other being cascaded geometry volumes. The data in all volumes are spherical functions, since they indicate how much light or blocking probability there is in all directions. Therefore, for the original method, every cell from all volumes contains a 2-band SH.

The CLPV is divided into four steps:

- obtain the indirect lighting sources and the blocking entities;
- inject the indirect light intensity and blocking probability into the volumes;
- propagate the indirect lighting in the volumes;
- use the volumes to illuminate the objects in your scenario;

The first step creates a reflective shadow map [Dachsbacher et. al. 2005] (RSM) for each light, which contains the indirect light sources and blocking entities, represented by their positions (depth value), directions (surface normal) and flux ($\varphi$). By using a deferred rendering pipeline, we can take advantage of the G-Buffer to get more blocking entities.

The second step injects SHs in the appropriate volume, one SH for each pixels in the RSM. Each SH represents the intensity of the indirect light sources, or the blocking probability of the rendered objects. The SH is modeled after a cosine lobe pointing to the same direction as the pixel normal. So, in this part with need the pre-computed cosine lobe coefficients (for SH or SRBF) and then rotate them.

The third step propagates the SHs only in the lighting volumes, since it makes no sense propagating the blocking probability. The propagation is iterative, meaning that the output of the previous iteration is the input of the next iteration, and the sum of all iteration's result represents the final indirect lighting. The SH in each cell is propagated to the cell's neighbors by following the main axes (positive and negative directions). For each neighbor we will add several SHs representing a cosine lobe, each of them pointing to one of the neighbor's cell faces, except the face in contact. The intensity of those SHs is based on the indirect lighting's intensity of the current cell, which is obtained by evaluating the current cell's SH in a particular direction. That direction starts from the neighbor face's center and ends in the current cell's center. Therefore, in this step, we need to evaluate SH or SRBF, create new SHs or SRBFs using the obtained value, and then rotate them.

The fourth and final step applies the indirect lighting into the objects contained in the volume. For each visible pixel, we get the appropriate SH from the smaller volume and we use the pixel's normal to get the indirect lighting's intensity. If the pixel uses a SH from the volume's boundary, we interpolate the current SH with the one from the next coarser volume. In this step we simply evaluate the SH or SRBF according to a particular direction.

## 3. Spherical Radial Basis Functions

From our brief explanation on how CLPV works (read the previous section), we determined that to integrate SRBF into CLPV we need two kinds of functions, one that evaluate the SRBF in a particular direction and another that rotates the SRBF's coefficients. In this section, we will not only explain how to use those functions, but we will also explain how the SRBF works and what kind of data we need to store to implement them in a shader program. Go check Appendix 2 for a possible implementation of the SRBF in a shader program.

The SRBFs consist on a set of coefficients and a single basis function, in which this function applies to a set of centers defined on the sphere's surface, and each coefficient is connected to only one center. As in the original method, these coefficients represents the indirect lighting and need to be stored in the LPV, while the basis function is needed to obtain the approximation of the original function, so we also need to store it. In the following equation, which represents an approximation of a function using a SRBF, the coefficients are represented by $c_i$, the basis function by $G$, and the centers by $\xi_i$:

$$f(\eta) \approx \sum_i c_i G(\eta \cdot \xi_i, \lambda_i)$$

As observed, the SRBF are very similar to SH, since the latter also comprises a set of coefficients associated with a set of basis functions (Figure 2). Besides the mentioned variables, we use $\eta$ as sampling point and there's still one unknown variable, $\lambda_i$. To explain the latter, we introduce the Abel-Poisson kernel as SRBF's basis function (Figure 2):

$$G^{Poisson}(\eta \cdot \xi, \lambda) = \frac{1 - \lambda^2}{[1 - 2\lambda(\eta \cdot \xi) + \lambda^2]^{3/2}}, 0 < \lambda < 1$$

This basis function is symmetric and its symmetry axis is represented by $\xi$ (the SRBF's centers) and $\lambda$ represents the opening of the symmetry (the point's distribution). For example, the opening is smaller when $\lambda$ is greater. This property makes SRBF more flexible than SH, since we can choose the more convenient distribution for the SRBF basis function, while the SH basis function has a fixed distribution. The basis function values can be pre-computed and stored into a buffer, and to ease our implementation, we use a fixed value for $\lambda$.

Now that we know how to obtain the approximation from the coefficients, let us explain how we can calculate those coefficients. We can obtain the final coefficients by normalizing intermediate ones, and the latter are obtained the same way as the SH's coefficients:

$$c_i = A^{-1} \alpha_i$$

$$\alpha_i = \int_{S^2} f(\eta) G(\eta \cdot \xi_i, \lambda_i) \, d\eta$$

As observed, we obtained the intermediate coefficients by integrating the original function and the basis function in a sphere. However, the basis function is not ortonormal, so we need to normalize the result, because if we did not, it would lead to overlapping values, since the basis function applied on one center does not know the result of the same basis function applied to another center. We use the matrix $A^{-1}$ to solve that problem, where each matrix components are represented by:

$$A_{ij} = H(\xi_i \cdot \xi_j, \lambda_i, \lambda_j)$$

The function $H$ is the singular integral, representing the convolution of two basis functions, allowing the basis function on one center to know the reach of the same basis function on another center:

$$H(\xi_i \cdot \xi_j, \lambda_i, \lambda_j) = (G_i *_2 G_j)(\xi_i \cdot \xi_j, \lambda_i, \lambda_j)$$
$$= \int_{S^2} G_i(\eta \cdot \xi_i, \lambda_i) G_j(\eta \cdot \xi_j, \lambda_j) d\omega(\eta)$$

$$H^{Poisson}(\xi_i \cdot \xi_j, \lambda_i, \lambda_j) = \frac{1 - (\lambda_i \lambda_j)^2}{\left[1 - 2(\lambda_i \lambda_j)(\xi_i \cdot \xi_j) + (\lambda_i \lambda_j)^2\right]^{3/2}},$$
$$0 < \lambda < 1$$

The values of the singular integral can be pre-computed and stored in a buffer for later use. Now that we know how to get the SRBF's coefficients, we will explain how to rotate a SRBF, since we need to rotate the LPV's virtual point lights. The SRBF rotation is invariant, so we can directly rotate the coefficients:

$$\tilde{f}(\eta) \approx \sum_i \tilde{c}_i G(\eta \cdot \xi_i, \lambda_i)$$

The rotation consists of one matrix operation, and unlike the 2-band SH, there is no workaround for this operation. Besides the matrix operation, we also need to normalize the results of the rotation:

$$\begin{bmatrix} \tilde{c} \end{bmatrix} = \begin{bmatrix} A^{-1} \end{bmatrix} \begin{bmatrix} R \end{bmatrix} \begin{bmatrix} c \end{bmatrix}$$
$$R_{ij} = H(\xi_i \cdot \tilde{\xi}_j, \lambda_i, \lambda_j)$$

In the above equation, $c$ represents the non-rotated coefficients, $R$ the rotation matrix (the convolution between the original centers and the rotated centers) and $A^{-1}$ he normalization matrix, the same we used to obtain the coefficients. The rotation matrix needs to be computed at runtime, but since it is based on the singular integral, all the matrix components can be obtained by sampling the singular integral buffer.

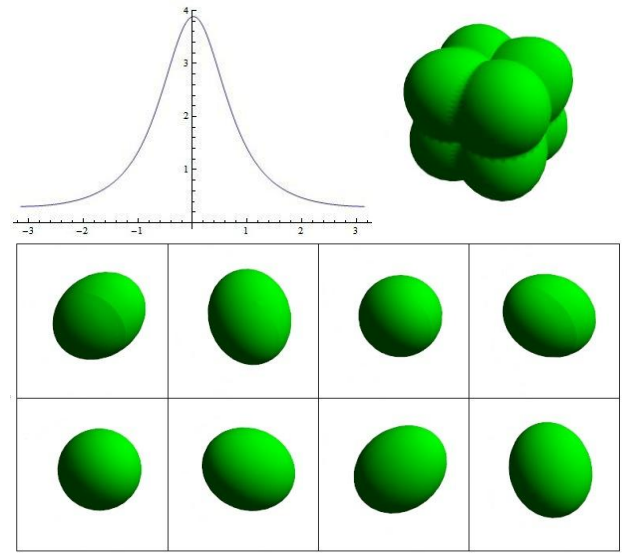

**Figure 2:** *Top left – Poisson kernel with $\lambda = 0.4$ ; Top right – Conjunction of the basis functions of a SRBF with 8 coefficients, using the Poisson kernel with $\lambda = 0.25$. Bottom – Same as the top right image, but separating the basis functions.*

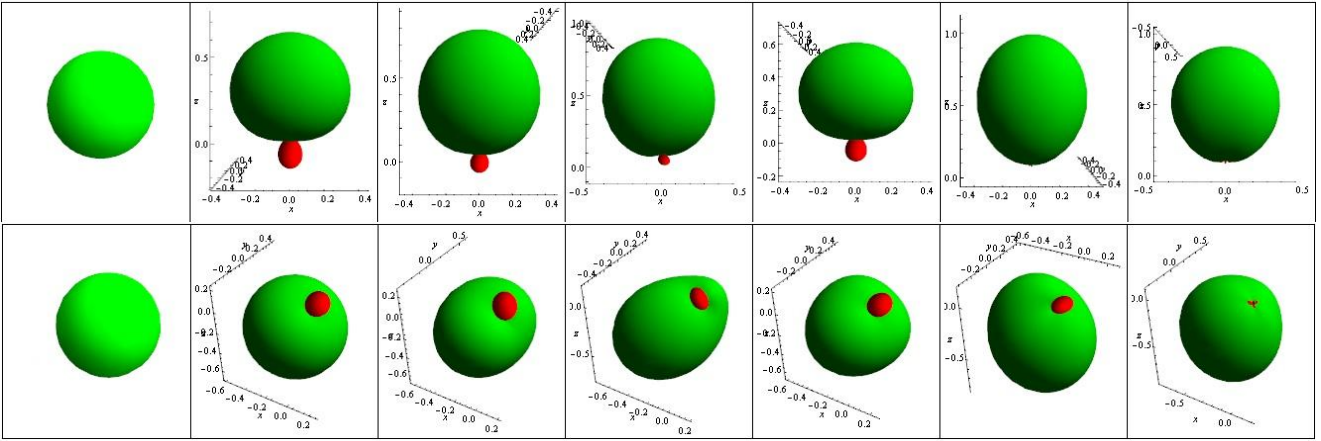

**Figure 3:** *From left to right: original function, 2-band SH, SRBF 4 coefficients, SRBF 8 coefficients version 1, 2 and 3, and SRBF 14 coefficients. On top we have a cosine lobe pointing to positive Z axis and on bottom a cosine lobe pointing to vector (-2, 1, -3).*

## 4. Qualitative evaluation

Considering that we are analyzing the approximation of spherical functions, we decided to assess the differences between the analytical representation of a spherical function, and its representation with a SH (2-band) and SRBFs (4, 8 and 14 coefficients). It should be noted that there are 3 versions of SRBFs with 8 coefficients:

- One with evenly spaced centers, which two of them are positioned onto the z-axis (version 1);

- One with uniformly spaced centers, each of them pointing to one of the cube's vertices. The centers of the cube's faces are placed on the main axes (version 2).

- One where each center is positioned onto the main axes, except two centers that are positioned onto arbitrary directions (version 3).

By using three versions of SRBFs with 8 coefficients, we will be able to see what effect the centers' disposition has on the representation.

The function in analysis is the cosine lobe, as it is the one used for the injection and propagation phase. We will analyze the approximation of that function in the positive z-axis (to simulate the propagation phase) and an arbitrary axis (to simulate the injection phase). The approximations in every main axis for the SH and all SRBFs are given in Appendix 1.

Let us begin with the cosine lobe on the positive z-axis. As ilustrated in Figure 3, it is very difficult to get a perfect representation of the cosine lobe, but this is perfectly normal since we only use a few coefficients to represent the original function. From all results, we found that the SRBF with 14 coefficients is the one that better represents the original function. We could also verify that the SRBFs with 4 and 8 coefficients (version 1 and 3) are very similar, but are worse than the SH's result. Finally, we also noticed that the SRBF with 8 coefficients (version 2) is very similar to the SH, indicating that the position of the SRBF's centers is extremely important for the approximation.

For the second test, the cosine lobe was pointing to an arbitrary direction (-2, 1, -3). As in the previous test, we found that the SRBF with 14 coefficients has the best representation, although it has some distortion at the origin (0, 0, 0). The second best representation goes to the SH. All other SRBFs are deformed, some more than others. The most deformed SRBFs are the ones with 4 and 8 coefficients (version 1 and 3), which confirms our conclusion in the previous test: the position of the SRBF's centers are extremely important for the approximation, especially when the SRBF uses a few coefficients.

## 5. Implementation Details

Our implementation of the CLPV is based on the implementation from [Kaplanyan et. al. 2010], so all the components used in the original method have the same implementation details. The implementation of the 2-band SH is based on the research from [Kirsch, A. 2010]. The light cascades and the geometry volume are 16-bit floating point RGBA 3D textures. We use 3 cascades for the light and 3 for the geometry, all of them with $32^3$ cells, where the smaller cascade has 12.5 meters, the second has 25 meters and the coarser has 50 meters. Our RSMs have a size of $256^2$ and contains a 32-bit depth buffer and 8-bit for normal and flux. The propagation process uses 8 iterations, but we tested our implementation with more iterations. Since we want to maintain the same performance requirements, our implementation use a 2-band SH and SRBFs with 4 and 8 coefficients per color channel for the LPV. We store the SRBF's basis function and singular integral in 32-bit texture buffer (one-dimensional textures). The rest of the needed variables (centers, normalization matrix and cosine lobe coefficients) are stored as constants in the GLSL shaders. We also decided not to simulate ambient lighting or other effects beyond the direct/indirect lighting and direct/indirect shadows.

## 6. Results and Discussion

To evaluate our implementation, we decided to use the same test scenario in [Kaplanyan et. al. 2010], the Sponza Atrium, since most surfaces of this model can only be lit indirectly. Besides that property, this model has many blocking objects (the atrium's columns) and high frequency surfaces (the atrium's curtains), allowing us to verify if the SRBF achieves a better result than the SH, like better shadow's definition or better representation of the original spherical function.

Our evaluation consists of two tests, one testing the global illumination in the Atrium, focusing in the curtain's lighting, and another testing the indirect shadows' definition. Both tests are made with our implementation of 2-band SH and the second version of the SRBF with 8 coefficients. The results of those tests are images (960x540 resolution), one for each simulation and another for the difference between the result of both implementations. The images that represent the difference between the implementations are exaggerated 15 times, but the indirect lighting intensity was not modified. Besides evaluating the simulation, we will also compare the performance of both implementations, using a computer with the following configuration: AMD Phenom II X4 955 (3.2Ghz), 4Gb RAM DDR3 (1666Mhz) and AMD Radeon HD6870 with 2Gb VRAM GDDR5.

The result of our first test (global illumination) is quite disappointing, since the SRBF implementation is slightly better at simulating the indirect lighting in small cascades, but worst when using bigger cascades (Figure 4). This is mainly due to the diversity of the indirect light sources, since bigger cascades means less detail in the RSM. Therefore, two neighbor pixels in the RSM can have completely different directions and intensity when the cascades have a huge size. This indicates that the number of coefficients for the SRBF is not enough to represent a huge diversity of cosine lobes, and while the result of SH might not be the most correct, it still gets better illumination, which is the main objective of the CLPV. We also tested our implementation with more iterations for the propagation step and/or with bigger cascades, and both results are disappointing. By increasing the number of iterations to 24, we also increased the error gathered in each iteration, which resulted in a bigger difference between the SH implementation and the SRBF implementation. By increasing the cascade's size, we verified that the SH suffers from the same problem and the difference between the two decrease when both uses bigger cascades. Still, the SH implementation gets better results.

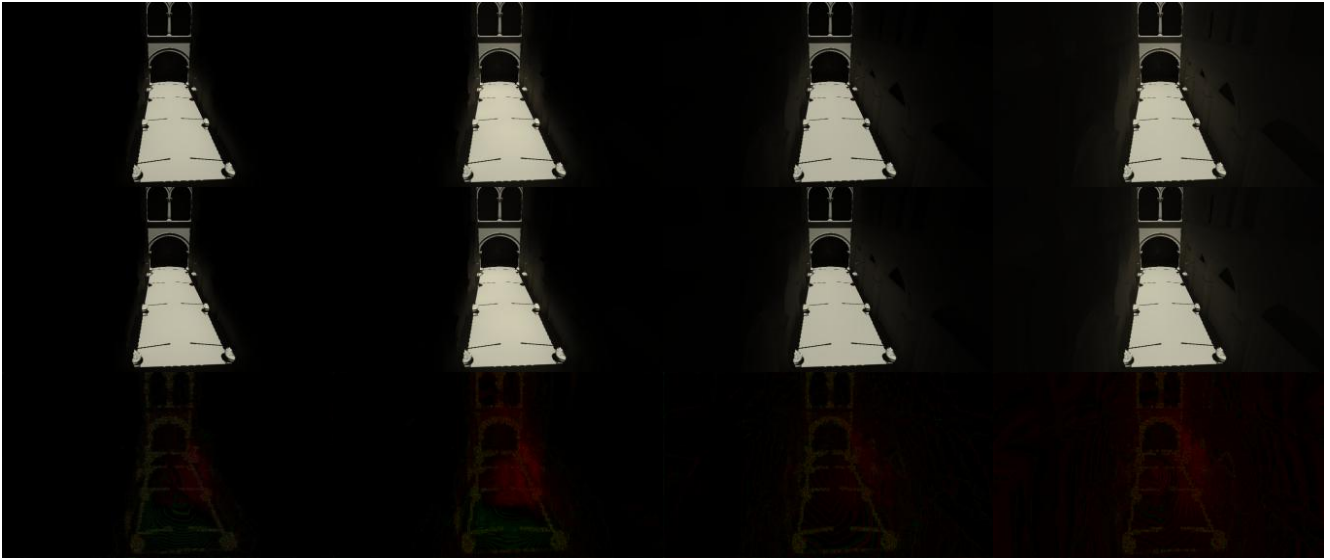

**Figure 4:** *Global illumination test with 2-band SH (top), SRBF with 8 coefficients version 2 (middle) and the difference between both (bottom). From left to right: normal cascade size with 8 iteration and 24 iterations, bigger cascade size (50 meters minimum) with 8 iterations and 24 iterations.*

The result of our second test (indirect shadow's definition) is even more disappointing, since the shadows on both SH and SRBF are too soft (Figure 5). However, given the settings we use, this result is expected. For example, we must not forget that the smallest cascade is 12.5 meters long, so each cell has about 40cm and consequently the pillars are 2 to 3 cells large. This number of cells is insufficient to create the blocking entities, especially if we consider the number of coefficients used for SH and SRBF. This number of cells is also insufficient to prevent the propagation of indirect light by the columns' sides. Although the shadows are not well defined, they are still simulated. Look at the shadows created by the first and second columns in Figure 5, which contrast with the light that passes through the curtain (which cannot block the indirect light because they have sufficient thickness).

|  | Injection (ms) Cascades small/medium/big | Propagation (ms) Cascades small/medium/big | Light Buffer (ms) |
|---|---|---|---|
| Results with indirect shadows ||||
| **SH2** | 2.028/ 2.292/ 2.339 | 2.241/ 2.565/ 2.275 | 1.086 |
| **SRFB4** | 2.090/ 2.314/ 2.351 | 4.139/ 4.974/ 4.200 | 1.214 |
| **SRBF8v2** | 3.544/ 4.488/ 4.299 | 9.663/10.602/9.386 | 1.723 |
| Results without indirect shadows ||||
| **SH2** | 2.025/ 2.294 / 2.340 | 1.957/ 2.125 / 1.986 | 1.135 |
| **SRBF4** | 2.086/ 2.281 / 2.344 | 3.183/ 3.797 / 3.177 | 1.206 |
| **SRBF8v2** | 3.511/ 4.460 / 4.276 | 7.427/ 7.794 / 7.019 | 1.721 |

**Table 1:** *Timings for the Sponza Atrium scene, using the configuration mentioned in the implementation details.*

## 7. Conclusion and Future Work

In the end, using SRBFs to simulate indirect lighting did not give us the results we desired. We expected to get a better result than SH (although only a slight improvement), while still getting an acceptable performance. Unfortunately, the result of the SRBF with eight coefficients (version 2) is practically equal to the SH of four coefficients, while getting worse performance (approximately three times slower). In addition to these bad results, we could also verify that changing the centers used in the SRBFs can result in unexpected behavior, indicating that it is necessary to test various centers in order to find the ideal solution to simulate indirect lighting in the given scenario.

Considering all of our research, we conclude the following: SRBFs with a few coefficients are not suitable to simulate correctly the spherical functions used to represent the indirect lighting, especially if we consider its unexpected behavior when we change the SRBF's centers.

We recommend that further investigations on CLPV can focus on increasing the coefficients for SH, and especially focusing on the use of GPGPU (OpenCL CUDA or DirectCompute) for all the calculations made in the injection, propagation and lighting phases. The latest will not only reduce the number of texture reads, but it will also allow the use of only one texture volume, even when the SH or SRBF use many coefficients.

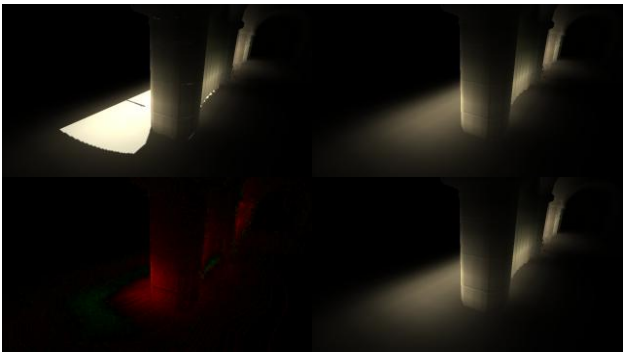

**Figure 5:** *Indirect shadow test with 2-band SH and direct lighting active (top left) and inactive (top right), with 8 coefficient SRBF (version 2) and direct lighting inactive (bottom right). Differences between both simulations without direct lighting (bottom left).*

# References

[Dachsbacher et. al. 2005] DACHSBACHER, C., AND STAMMINGER, M., 2005. Reflective shadow maps. In Proc. of the Symposium on Interactive 3D Graphics and Games.

[Figueiredo et al. 2010] FIGUEIREDO, M., OLIVEIRA, J., ARAUJO, B. AND PEREIRA J., 2010. An Efficient Collision Detection for Point Cloud Models. In Proceedings of the 20th International Conference on Computer Graphics and Vision; Russia, pp. 30-37, 2010.

[Kaplanyan et. al. 2010] KAPLANYAN, A., AND CARSTEN D., 2010. Cascaded light propagation volumes for real-time indirect illumination. In I3D '10: Proceedings of the ACM SIGGRAPH Symposium on Interactive 3D Graphics and Games.

[Kirsch, A. 2010] KIRSCH, A. 2010. Light Propagation Volumes - Annotations. In http://blog.blackhc.net/wp-content/uploads/2010/07/lpv-annotations.pdf

[Ritschel et. al. 2009] RITSCHEL, T., GROSCH, T., AND SEIDEL, H.-P., 2009. Approximating dynamic global illumination in image space. In I3D '09 : Proceedings of the 2009 symposium on Interactive 3D graphics and games.

[Sloan et. al. 2002] SLOAN, P.-P., KAUTZ, J., AND SNYDER, J. 2002. Precomputed radiance transfer for real-time rendering in dynamic, low-frequency lighting environments. In ACM Transactions on Graphics (Proc. of SIGGRAPH 2002) 21.

[Tsai et. al. 2006] TSAI, Y., SHIH Z. 2006. All-frequency precomputed radiance transfer using spherical radial basis functions and clustered tensor approximation. In ACM Transactions on Graphics (Proc. of SIGGRAPH 2006).

# Appendix 1

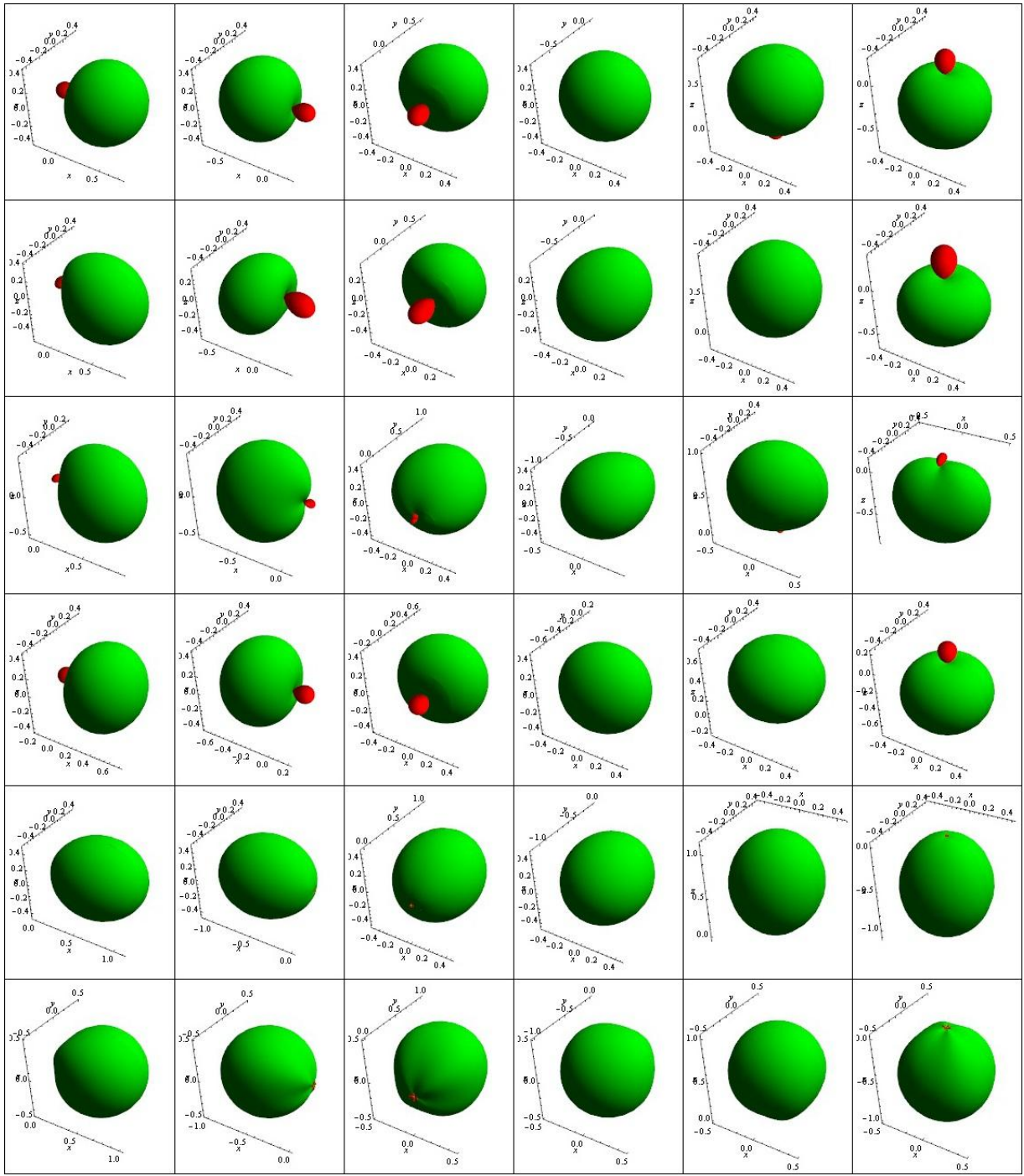

**Figure 6:** *Cosine lobes pointing to the main axes (left to right: X, -X, Y, -Y, Z, -Z), using SH and SRBFs (top to bottom: 2-band SH, SRBF with 4 coefficients, 8 coefficients version 1, version 2, version 3 and SRBF with 14 coefficients).*

# Appendix 2

Functions used to rotate and evaluate a SRBF with 8 coefficients into a particular direction:

```
vec8 srbfRotateCosineLobe( vec3 direction ) {
    direction = normalize( direction );
    mat3 rotMatrix = getRotationMatrix( vec3( 0.0, 0.0, 1.0 ), direction );
    vec3 rotCenter0 = rotMatrix * center0;
    vec3 rotCenter1 = rotMatrix * center1;
    vec3 rotCenter2 = rotMatrix * center2;
    vec3 rotCenter3 = rotMatrix * center3;
    vec3 rotCenter4 = rotMatrix * center4;
    vec3 rotCenter5 = rotMatrix * center5;
    vec3 rotCenter6 = rotMatrix * center6;
    vec3 rotCenter7 = rotMatrix * center7;

    int texSize = textureSize( uPoissonSingularIntegral ) - 1;

    float e00 = texelFetch( uPoissonSingularIntegral, int(( dot(center0, rotCenter0) + 1 ) * 0.5 * texSize) ).r;
    float e01 = texelFetch( uPoissonSingularIntegral, int(( dot(center0, rotCenter1) + 1 ) * 0.5 * texSize) ).r;
    float e02 = texelFetch( uPoissonSingularIntegral, int(( dot(center0, rotCenter2) + 1 ) * 0.5 * texSize) ).r;
    float e03 = texelFetch( uPoissonSingularIntegral, int(( dot(center0, rotCenter3) + 1 ) * 0.5 * texSize) ).r;
    float e04 = texelFetch( uPoissonSingularIntegral, int(( dot(center0, rotCenter4) + 1 ) * 0.5 * texSize) ).r;
    float e05 = texelFetch( uPoissonSingularIntegral, int(( dot(center0, rotCenter5) + 1 ) * 0.5 * texSize) ).r;
    float e06 = texelFetch( uPoissonSingularIntegral, int(( dot(center0, rotCenter6) + 1 ) * 0.5 * texSize) ).r;
    float e07 = texelFetch( uPoissonSingularIntegral, int(( dot(center0, rotCenter7) + 1 ) * 0.5 * texSize) ).r;
    float e10 = texelFetch( uPoissonSingularIntegral, int(( dot(center1, rotCenter0) + 1 ) * 0.5 * texSize) ).r;
    float e11 = texelFetch( uPoissonSingularIntegral, int(( dot(center1, rotCenter1) + 1 ) * 0.5 * texSize) ).r;
    .
    .
    .
    float e77 = texelFetch( uPoissonSingularIntegral, int(( dot(center7, rotCenter7) + 1 ) * 0.5 * texSize) ).r;

    mat8 srbf8RotMatrix = mat8(
        vec4( e00, e01, e02, e03 ), vec4( e04, e05, e06, e07 ),
        vec4( e10, e11, e12, e13 ), vec4( e14, e15, e16, e17 ),
        vec4( e20, e21, e22, e23 ), vec4( e24, e25, e26, e27 ),
        vec4( e30, e31, e32, e33 ), vec4( e34, e35, e36, e37 ),
        vec4( e40, e41, e42, e43 ), vec4( e44, e45, e46, e47 ),
        vec4( e50, e51, e52, e53 ), vec4( e54, e55, e56, e57 ),
        vec4( e60, e61, e62, e63 ), vec4( e64, e65, e66, e67 ),
        vec4( e70, e71, e72, e73 ), vec4( e74, e75, e76, e77 )
    );

    return rotateVec8( srbf8NormMatrix, rotateVec8( srbf8RotMatrix, srbf8CosLobeC ) );
}

vec8 srbfEvaluateBasisFunctions( vec3 direction ) {
    int texSize = textureSize( uPoissonKernel ) - 1;
    float c0 = texelFetch( uPoissonKernel, int( ( dot( center0, direction ) + 1 ) * 0.5 * texSize ) ).r;
    float c1 = texelFetch( uPoissonKernel, int( ( dot( center1, direction ) + 1 ) * 0.5 * texSize ) ).r;
    float c2 = texelFetch( uPoissonKernel, int( ( dot( center2, direction ) + 1 ) * 0.5 * texSize ) ).r;
    float c3 = texelFetch( uPoissonKernel, int( ( dot( center3, direction ) + 1 ) * 0.5 * texSize ) ).r;
    float c4 = texelFetch( uPoissonKernel, int( ( dot( center4, direction ) + 1 ) * 0.5 * texSize ) ).r;
    float c5 = texelFetch( uPoissonKernel, int( ( dot( center5, direction ) + 1 ) * 0.5 * texSize ) ).r;
    float c6 = texelFetch( uPoissonKernel, int( ( dot( center6, direction ) + 1 ) * 0.5 * texSize ) ).r;
    float c7 = texelFetch( uPoissonKernel, int( ( dot( center7, direction ) + 1 ) * 0.5 * texSize ) ).r;

    return vec8( vec4( c0, c1, c2, c3 ), vec4( c4, c5, c6, c7 ) );
}
```